\documentclass{article}
\usepackage{spconf,amsmath,graphicx}
\usepackage[skip=0pt]{caption}

\title{Confidence guided depth completion network}
\name{Yongjin Lee \qquad Seokjun Park \qquad Beomgu Kang \qquad Hyunwook Park}
\address{Korea Advanced Institute of Science and Technology \\ Daejeon, Republic of Korea}
\begin{document}
%
\maketitle
\begin{abstract}
The paper proposes an image-guided depth completion method to estimate accurate dense depth maps with fast computation time. The proposed network has two-stage structure. The first stage predicts a first depth map. Then, the second stage further refines the first depth map using the confidence maps. The second stage consists of two layers, each of which focuses on different regions and generates a refined depth map and a confidence map. The final depth map is obtained by combining two depth maps from the second stage using the corresponding confidence maps. Compared with the top-ranked models on the KITTI depth completion online leaderboard, the proposed model shows much faster computation time and competitive performance.
\end{abstract}

\section{Introduction}
\label{sec:intro}
Depth information is important in computer vision for various applications such as autonomous driving, and 3D reconstruction. For depth measurement, Light Detection and Ranging (LiDAR) sensors are commonly used, which measure accurate depth information. However, the LiDAR sensor can provide the limited amount of valid depth points due to the hardware limitations. For example, the projected depth map of point cloud data measured by the Velodyne HDL-64E has a density of approximately 4\%, which is insufficient for high-level applications such as autonomous driving \cite{uhrig2017sparsity}.

To overcome the issue of the data sparsity, image-guided depth completion methods have been recently investigated. The objective of the image-guided depth completion is to predict accurate depth maps by effectively utilizing the guidance images. Multi-branch networks adopting confidence maps to obtain final depth maps were proposed, and showed high performance improvement \cite{van2019sparse,qiu2019deeplidar}. However, the color image was simply concatenated with the sparse depth map and, therefore a more effective method for utilizing guidance images was necessary. Tang \textit{et al.} \cite{tang2020learning} proposed content-dependent and spatially-variant kernels, which were generated from color images, to extract feature maps from depth maps. Liu \textit{et al.} \cite{liu2020fcfr} proposed two-stage frameworks of sparse-to-dense and coarse-to-fine and further improved the performance. In the coarse-to-fine stage, the features of the color image and the coarse depth map were combined by channel shuffle and energy-based fusion. Cheng \textit{et al.} \cite{cheng2019learning} proposed convolutional spatial propagation network (CSPN) to learn the affinity among neighboring pixels. The edge-preserving refinement was performed with the pixel-wise operation.

Although these methods showed a high degree of performance improvement, the characteristics of the depth map was not fully explored. Since 3D point cloud data is projected on the 2D images, the 2D depth map shares the property of extremely unbalanced distribution of structures in image space resulted from the perspective projection. Near objects have a large area in the image plane with sufficient depth points, whereas distant objects have a small area with insufficient depth points. In autonomous driving datasets, most of the depth data is distributed within a distance of 20 meters, and the variance of depth for distant object farther than 30 meters is quite large \cite{li2020multi}. Recently, the unbalanced data distribution was taken into consideration for the effective fusion of two different types of data. Multi-scale cascade hourglass network was proposed, which predicted depth maps of different sizes to represent the different data distributions \cite{li2020multi}. Lee \textit{et al.} \cite{lee2021depth} changed the regression task to the classification task by separating the depth map along the channel axis with intervals. These methods considered the data distribution of depth maps, but consumed a lot of computational resources. High computational cost remains to be a challenge for real-time applications such as autonomous driving, which has a limited computational resource.

In this paper, we propose an efficient and effective two-stage depth completion network, which predicts a dense coarse depth map in the first stage and refines it in the second stage. The shallow feature fusion module (SFFM) is applied to combine the depth map and the color image in both stages. The second stage consists of color-refinement (CR) layer and depth refinement (DR) layer, each of which predicts the depth map and the confidence map. The final depth map is obtained by combining two depth maps from the refinement layers using the corresponding confidence maps. The proposed model provides accurate depth map using the confidence maps. Therefore, the proposed model provides accurate depth map with the fast runtime.

\begin{figure*}[!t]
\centering
\includegraphics[width=1\linewidth]{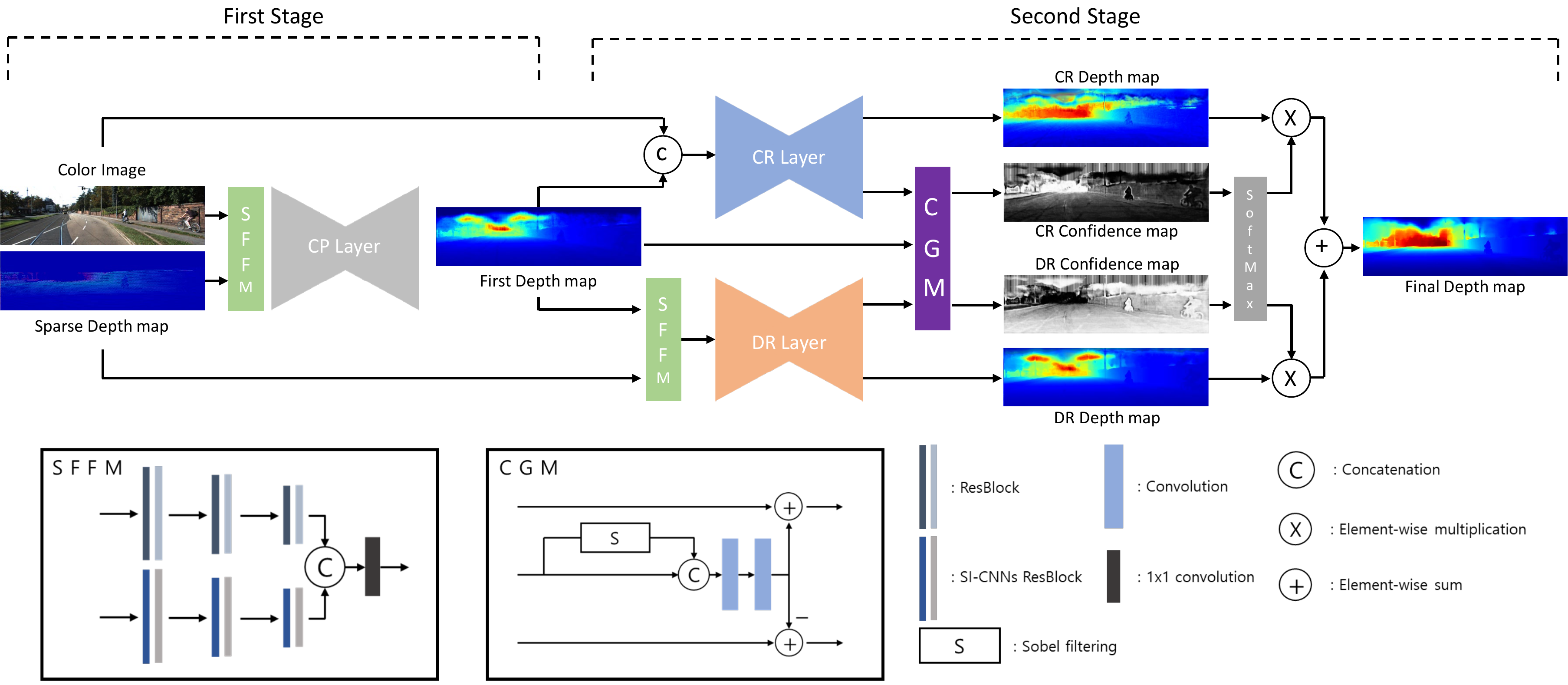}
\hfil
\caption{\textbf{Overall diagram of the proposed two-stage model.} The coarse dense depth map is obtained in the first stage by coarse-prediction (CP) layer, and refined by color-refinement (CR) layer and depth-refinement (DR) layer in second stage. Shallow feature fusion module (SFFM) is applied to fusion of the sparse depth map and dense data}
\label{fig:architecture}
\end{figure*}

\section{METHODOLOGY}
\subsection{Network}
The entire architecture of the proposed model is described in Fig. \ref{fig:architecture}. Note that all encoder-decoder blocks contain residual blocks and decoders. Our model is a two-stage network. In the first stage, a coarse dense depth map called first depth map is predicted from a color image and a sparse depth map as follows:
\begin{equation}
    D_{c} = CP(SFFM(I_{c}, I{s}))
\end{equation}
where $D_{c}$ denotes the coarse dense depth map from the first stage, $I_{c}$ denotes the input color image, $I_{s}$ denotes the input sparse depth map, the CP is the coarse-prediction layer in Fig. \ref{fig:architecture}, and the SFFM is the proposed feature fusion module.

The SFFM extracts the features that is robust to the depth validity. The color images are processed with the conventional convolution, whereas the sparse depth maps are processed with the sparsity invariant CNNs (SI-Conv \cite{uhrig2017sparsity}). The SI-Conv is an effective method for processing sparse data because it helps the kernel circumvent the effect of irregular validity of sparse data. Therefore, the SI-Conv increases the feature density from the sparse depth map. 1$\times$1 convolution is also adopted to combine the concatenated features of dense color image and semi-dense sparse depth map, enabling the construction of completely-dense feature map. Since the 1$\times$1 convolution is operated in a pixel-wise manner, it greatly simplifies the fusion regardless of the validity of neighboring pixels.

\begin{figure}[!t]
\centering
\includegraphics[width=1\linewidth]{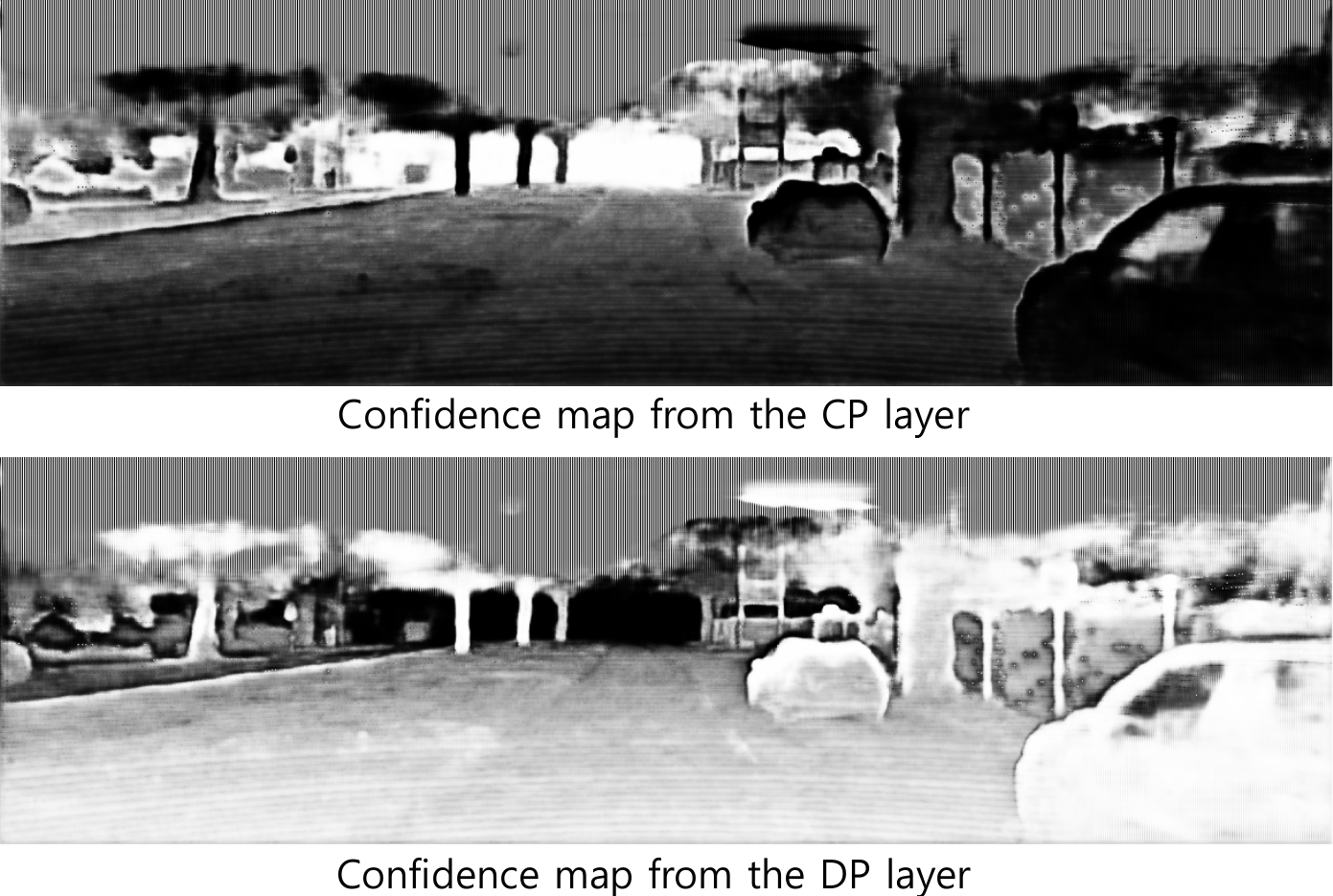}
\hfil
\caption{The proposed confidence maps from the color-refinement (CR) layer and the depth-refinement (DR) layer}
\label{fig:confidence_maps}
\end{figure}

In the second stage, the coarse dense depth map is refined by the color-refinement (CR) layer with the color image and the depth-refinement (DR) layer with the sparse depth map, which can be written as:
\begin{flalign}
&D_{cr},\ C_{cr} = CR(D_{c},I_{c}) \\
&D_{dr},\ C_{dr} = DR(SFFM(D_{c},I_{s}))
\end{flalign}
where $D_{cr}$ denotes the dense depth map from the CR layer, $C_{cr}$ denotes the confidence map from the CR layer, $D_{dr}$ denotes the dense depth map from the DR layer, $C_{dr}$ denotes the confidence map from the DR layer.

The CR and DR layers do not need to predict accurate depth maps for all regions. Each layer refines the exclusive region that can effectively make use of the distinctive characteristics of different input data spaces. The two layers are complementary, and the final depth map is obtained through the fusion of the depth maps using the confidence map, which can be written as:
\begin{equation}
    D_{f}(u,v) = \frac{e^{C_{cr}(u,v)} \cdot D_{cr}(u,v) + e^{C_{dr}(u,v)} \cdot D_{dr}(u,v) }{e^{C_{cr}(u,v)} + e^{C_{dr}(u,v)}}
\end{equation}
where $(u,v)$ denotes a pixel, and $D_{f}$ denotes the final depth map.

In addition, we design the confidence guidance module (CGM) to estimate the reliable confidence maps, which do not have the ground truths. Each layer should address the desirable region adequately that the CR layer accounts for the object boundaries and far distance objects, and the DR layer accounts for the inner region and the close distance objects. In particular, the Sobel filter \cite{kanopoulos1988design} is used to estimate the boundaries of objects. From the CGM, the CR layer increases the confidence for object boundary and far distance, and the DR layer lowers the confidence at that regions, which can be written as:
\begin{equation}
    C'_{cr},\ C'_{dr} = CGM(D_{c}, C_{cr}, C_{dr})
\end{equation}
where $C'_{cr}$ denotes the CR confidence map, and $C'_{dr}$ denotes the DR confidence map. Fig. \ref{fig:confidence_maps} shows an example of the CR and DR confidence maps.

Therefore, the final depth map can be rewritten as:
\begin{equation}
    D'_{f}(u,v) = \frac{e^{C'_{cr}(u,v)} \cdot D_{cr}(u,v) + e^{C'_{dr}(u,v)} \cdot D_{dr}(u,v) }{e^{C'_{cr}(u,v)} + e^{C'_{dr}(u,v)}}
\end{equation}
where $D'_{f}$ denotes the final depth map of the proposed model.

\subsection{Loss Function}
The ground truth depth map is semi-dense and invalid pixels are represented as 0. Therefore, the loss is defined only for the valid pixels by calculating the mean squared error (MSE) between the final depth map and the ground truth map as follow:
\begin{equation}
    L_{final} = \frac{1}{|V|} \sum_{(u,v)\in V} \left\| (D_{gt}(u,v) - D'_{f}(u,v)) \right\|^{2}
\end{equation}
where $V$ denotes the set of valid pixels, $D'_{f}(u,v)$ denotes the final depth map at pixel $(u,v)$ and $D_{gt}(u,v)$ denotes the ground truth depth map at pixel $(u,v)$. 

To train the network more stable, the loss for the first stage depth map was also computed in the early epochs as follows:
\begin{equation}
    L_{first} = \frac{1}{|V|} \sum_{(u,v)\in V} \left\| (D_{gt}(u,v) - D_{c}(u,v)) \right\|^{2}
\end{equation}
where $D_{c}(u,v)$ denotes the coarse depth map called first depth map at pixel $(u,v)$.

The overall loss can be written as:
\begin{equation} 
    L_{total} = C_{first} \times L_{first} + L_{final}
\end{equation}
where $C_{first}$ is a hyper-parameter of 0.3 at the first epoch and reduces to 0 at 5$^{th}$ epoch

\begin{figure*}[!t]
\centering
\includegraphics[width=1\linewidth]{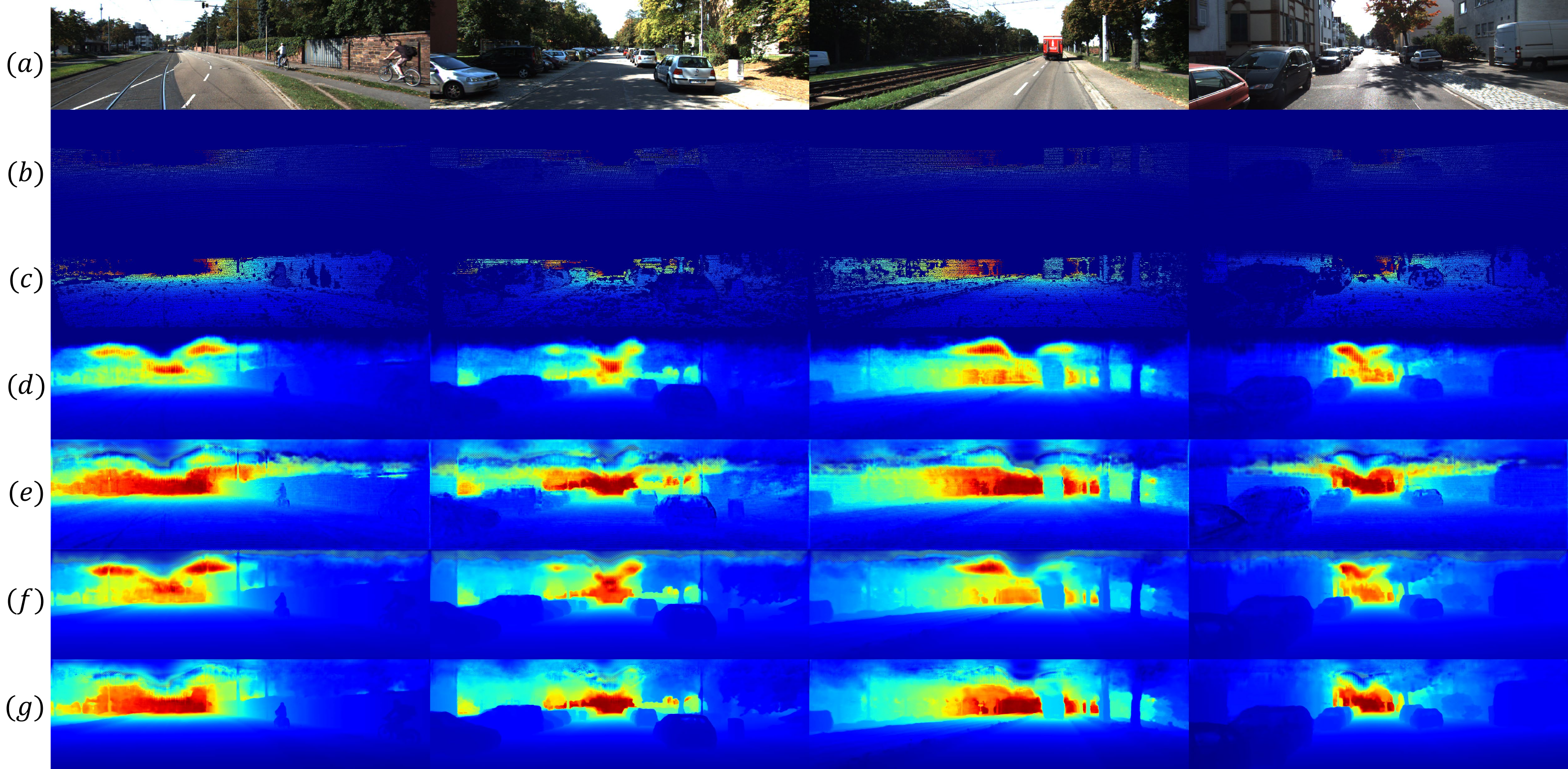}
\hfil
\caption{\textbf{Depth completion results on the KITTI depth completion validation dataset.} (a) color images, (b) sparse depth maps, (c) ground truth depth maps, (d) first depth maps from CP layer, (e) CR depth maps from the CR layer, (f) DR depth maps from the DR layer, and (g) final depth maps.}
\label{fig:result}
\end{figure*}

\section{EXPERIMENTS}
\subsection{Experimental setup}
\textbf{KITTI depth completion dataset:} The KITTI depth completion dataset is a large real-world street view dataset captured for autonomous driving research \cite{uhrig2017sparsity}, \cite{geiger2013vision}. It provides sparse depth maps of 3D point cloud data and corresponding color images. The sparse depth maps have a valid pixel density of approximately 4\% and the ground truth depth maps have a density of 16\% compared to the color images (\cite{uhrig2017sparsity}). This dataset contains 86K training set, 1K validation set, and 1K test set without ground truth. KITTI receives the predicted depth maps of the test set and provides the evaluation results.

\noindent
\textbf{Implementation details:} We trained our network on two NVIDIA TITAN RTX GPUs with batch size of 8 for 25 epochs. We used the ADAM optimizer \cite{kingma2014adam} with $\beta_{1} = 0.9, \beta_{2} = 0.99$ and the weight decay of $10^{-6}$. The learning rate started at 0.001 and was halved for every 5 epochs. For data augmentation, color jittering and horizontal random flipping were adopted.

\noindent
\textbf{Evaluation metrics:} We adopt commonly used metrics for comparison study, including the inverse root mean squared error (iRMSE [1/km]), the inverse mean absolute error (iMAE [1/km]), the root mean squared error (RMSE [mm]), the mean absolute error (MAE [mm]) and the runtime ([s]).

\subsection{Comparison with state-of-the-art methods}
We evaluated the proposed model on the KITTI depth completion test set. Table \ref{table:compare} shows the comparison results with other top ranked methods. The proposed model shows the fastest runtime, and shows comparable performance against SoTA model PENet \cite{hu2021penet} and higher than other top-ranked methods on RMSE, which is the most important metric in depth completion. Moreover, our model shows higher performance than SoTA model in iRMSE and iMAE.

\begin{table}[!t]
\centering
\resizebox{0.5\textwidth}{!}{
\begin{tabular}{l|ccccc}
\hline
Method                              & iRMSE          & iMAE          & RMSE            & MAE             & Runtime \\ \hline
CrossGuidance \cite{lee2020deep}    & 2.73           & 1.33          & 807.42          & 253.98          & 0.2 s   \\
PwP \cite{xu2019depth}              & 2.42           & 1.13          & 777.05          & 235.17          & 0.1 s   \\
DeepLiDAR \cite{qiu2019deeplidar}   & 2.56           & 1.15          & 758.38          & 226.50          & 0.07s   \\
CSPN++ \cite{cheng2020cspn++}       & \textbf{2.07}  & \textbf{0.90} & 743.69          & 209.28          & 0.2 s   \\
ACMNet \cite{zhao2021adaptive}      & 2.08           & 0.90          & 744.91          & \textbf{206.09} & 0.08 s  \\
GuideNet \cite{tang2020learning}    & 2.25           & 0.99          & 736.24          & 218.83          & 0.14 s  \\
FCFR-Net \cite{liu2020fcfr}         & 2.20           & 0.98          & 735.81          & 217.15          & 0.13 s  \\
PENet \cite{hu2021penet}            & 2.17           & 0.94          & \textbf{730.08} & 210.55          & 0.032s  \\ \hline
Ours                                & 2.11           & 0.92          & 733.69          & 211.15          & \textbf{0.015 s}  \\ \hline
\end{tabular}
}
\caption{Comparison with state-of-the-art methods on the KITTI Depth Completion test set.}
\label{table:compare}
\end{table}

\subsection{Ablation studies}
In this section, we conducted ablation studies on the KITTI validation dataset to verify the effectiveness of the proposed model. The experimental results are shown in Table \ref{table:Ablation}. B is a basic two-stage model, which predicts a first depth map from the concatenated input of a color image and a sparse depth map in first stage, and predicts a final depth map from the concatenated input of a first depth map, a color image and a sparse depth map. The CR and DR replace the second stage of the basic two-stage model. Each encoder-decoder takes a first depth map concatenated with a color image or a sparse depth map. The results show the CR and DR layers archives significant improvement in all the metrics, and both of the SFFM and the CGM also gives a performance improvement. 

\section{CONCLUSION}
The paper proposes an image-guided depth completion method to estimate accurate dense depth maps with fast computation time. The proposed network has two-stage structure including a shallow feature fusion module (SFFM), a coasre-prediction (CP) layer, color-refinement (CR) and depth-refinement (DR) layers, and confidence guidance module (CGM). The first depth map from the CP layer is effectively refined in the CR and DR layers and the two depth maps are effectively combined with the confidence map from the CGM. Compared with the top-ranked models on the KITTI depth completion online leaderboard, the proposed model shows much faster computation time and competitive performance.

\begin{table}[!t]
\centering
\resizebox{0.5\textwidth}{!}{
\begin{tabular}{l|cccc}
\hline
\multicolumn{1}{c|}{Models} & iRMSE          & iMAE          & RMSE            & MAE    \\ \hline
B                           & 2.29           & 0.98          & 779.68          & 224.91 \\
CR and DR                   & 2.17           & 0.93          & 769.28          & 213.30 \\
CR and DR + SFFM            & 2.17           & 0.91          & 764.93          & 212.71 \\ \hline
CR and DR + SFFM + CGM       & \textbf{2.17}  & \textbf{0.91} & \textbf{759.90} & \textbf{209.25}
\end{tabular}
}
\caption{Ablation studies on the KITTI depth completion validation set. B: basic two-stage model, CR and DR: the second stage of B is replaced with the CR and DR layers}
\label{table:Ablation}
\end{table}

\section{ACKNOWLEDGEMENT}
This work was conducted by Center for Applied Research in Artificial Intelligence(CARAI) grant funded by Defense Acquisition Program Administration(DAPA) and Agency for Defense Development(ADD) (UD190031RD).

\newpage

\bibliographystyle{IEEEbib}
\bibliography{strings,refs}

\end{document}